\documentclass{article}

\usepackage{arxiv}
\usepackage[utf8]{inputenc} % allow utf-8 input
\usepackage[T1]{fontenc}    % use 8-bit T1 fonts
\usepackage[colorlinks=true]{hyperref}       % hyperlinks
\hypersetup{allcolors=black}
\usepackage{url}            % simple URL typesetting
\usepackage{booktabs}       % professional-quality tables
\usepackage{amsfonts}       % blackboard math symbols
\usepackage{eqnarray,amsmath} % Special equations
\usepackage{nicefrac}       % compact symbols for 1/2, etc.
\usepackage{microtype}      % microtypography
\usepackage{lipsum}		% Can be removed after putting your text content
\usepackage{graphicx}
\usepackage{cleveref}
\usepackage[caption = false]{subfig}
\usepackage{float}
\usepackage{xurl}
\usepackage{doi}
\usepackage{natbib}
\setcitestyle{authoryear,open={(},close={)}} %Citation-related commands
\usepackage{textcomp}  % for € etc.
\usepackage{eurosym}

\usepackage{xr}
\makeatletter
\newcommand*{\addFileDependency}[1]{% argument=file name and extension
\typeout{(#1)}
\@addtofilelist{#1}
\IfFileExists{#1}{}{\typeout{No file #1.}}
}
\makeatother
\usepackage{color}
\renewcommand*{\vec}[1]{\ensuremath{\mathrm{#1}}}

\newcommand{\difftime}{\ensuremath{\tau}}

\newcommand{\rX}{\vec{x}}
\newcommand{\rY}{\vec{y}}

\newcommand{\hod}{\ensuremath{_{\mathrm{HOD}}}}
\newcommand{\mnth}{\ensuremath{_{\mathrm{M}}}}

\usepackage[acronym]{glossaries}
\newacronym{cf}{CF}{Capacity Factor}

\title{Assessing the risk of future Dunkelflaute events for Germany using generative deep learning}
\renewcommand{\shorttitle}{}

\date{\today} 					% Or removing it
\author{
    Felix M. Strnad \\
    University of Tübingen, Tübingen AI Center, Germany\\
    \AND
    Jonathan Schmidt \\
    University of Tübingen, Tübingen AI Center, Germany\\
    \AND
    Fabian Mockert \\
    Karlsruhe Institute of Technology, Institute of Statistics, Germany\\
    \AND
    Philipp Hennig \\
    University of Tübingen, Tübingen AI Center, Germany\\
    \AND
    Nicole Ludwig\\
    University of Tübingen, Tübingen AI Center, Germany\\
}% Uncomment to remove the date
\hypersetup{
pdftitle={Synchronization},
pdfauthor={Felix Strnad},
pdfkeywords={},
}
\usepackage{verbatim}

\newcommand{\detailtexcount}[1]{%
  \immediate\write18{texcount -merge -sum -q #1.tex > #1.wcdetail }%
  \verbatiminput{#1.wcdetail}%
}

\usepackage{lineno}
\usepackage{setspace}
\begin{document}
%TC:ignore
\detailtexcount{article}
\newpage

\maketitle

% Maximum of 300 words
\begin{abstract}
  The European electricity power grid is transitioning towards renewable energy sources, characterized by an increasing share of off- and onshore wind and solar power.
  However, the weather dependency of these energy sources poses a challenge to grid stability, with so-called Dunkelflaute events--- periods of low wind and solar power generation---being of particular concern due to their potential to cause electricity supply shortages.
  In this study, we investigate the impact of these events on the German electricity production in the years and decades to come.
  For this purpose, we adapt a recently developed generative deep learning framework to downscale climate simulations from the CMIP6 ensemble.
  We first compare their statistics to the historical record taken from ERA5 data. Next, we use these downscaled simulations to assess plausible future occurrences of Dunkelflaute events in Germany under the optimistic low (SSP2-4.5) and high (SSP5-8.5) emission scenarios.

  Our analysis indicates that both the frequency and duration of Dunkelflaute events in Germany in the ensemble mean are projected to remain largely unchanged compared to the historical period. This suggests that, under the considered climate scenarios, the associated risk is expected to remain stable throughout the century.

\end{abstract}

%TC:endignore
\newpage

% Text: Maximum of 4000 words, excluding figures, tables, captions references acknowledgements, headings.

\section{Introduction}
Meeting the EU’s Green Deal targets (net-zero for emissions by 2050; $\geq55\%$ reduction by 2030) demands a rapid expansion of variable renewables, especially wind and solar \citep{EuropeanCommission2019GreenDeal}.
Renewables already supplied 44.7\% of EU electricity in 2023 and 62.7\% in Germany, where wind and solar accounted for 31.5\% and 13.8\%, respectively \citep{Eurostat2024RenewablesLead2023, Collins2025Germany, Enerdata2025GermanyWindSolar}.
This trajectory is expected to continue, with the share of renewables projected to reach up to 80\% by 2030 \citep{Meza2025AccelerateRenewables}.
As weather-dependent sources grow, the power grid becomes increasingly sensitive to meteorological variability.
In particular, simultaneous lulls in both wind and solar output pose a serious challenge. These so-called ``Dunkelflaute'' (or ``dark doldrums'') events, typically occurring in winter under prolonged high-pressure, cloudy, low-wind conditions, can knock out most renewable generation at once.
Germany experienced pronounced examples in November and December 2024: a week-long event in November cut renewables to 30\%, of supply and drove baseload prices to 145\,\euro/MWh (compared to $\sim$40\,\euro/MWh on average) \citep{TimeraEnergy2025}, while a December episode briefly pushed prices past 175\,\euro/MWh \citep{TimeraEnergy2025}. These episodes demonstrate that Dunkelflaute events already cause acute supply shortfalls and extreme price volatility, underscoring the need to assess their future risk under climate change.
Prior work has mainly characterized Dunkelflaute events under historical climate but not projected them into the future. For example, meteorological studies of Germany show that they occur mostly in winter when solar irradiance is naturally low and winds---normally strong in this season---become stagnant; they last a few days and typically recur a couple of times per winter \citep{Mockert2023}.
Similarly, \citet{Kittel2024b} examined historical events across Europe using aggregated time series per country.
There is a growing number of studies highlighting how wind and solar lulls combine \citep{Kittel2024, Biewald2025}, but they focus on past data and fixed portfolio effects, and thus do not necessarily allow conclusions about future scenarios under climate change.

To assess future Dunkelflaute risk, climate projections with high spatial and temporal resolution are needed to capture local weather patterns and short-term variability.
Future climate projections from output of global climate model (GCM) simulations  are provided by the Coupled Model Intercomparison Project Phase 6 (CMIP6) extending to the year 2100 \citep{Eyring2016}.
However, their native resolution, typically on the order of 100\,km, remains too coarse for detailed, local impact assessments. Downscaling therefore refines GCM output either dynamically or statistically.
Dynamical downscaling, on the one hand, uses high-resolution regional climate models (RCMs) driven by GCM output to simulate local climate processes explicitly \citep{Giorgi2009}.
Databases, such as EURO-CORDEX, provide these high-resolution climate projections (12-50 km) for Europe \citep{Kotlarski2014}.
There are first studies that applied such dynamical downscaling approaches to CMIP6 data to investigate future Dunkelflaute events \citep{Duchene2024}.
Data from the EURO-CORDEX database is used to derive energy relevant variables by the Copernicus Pan-European Climate Database (PECD v4.2) \citep{C3S_PECD_v4}, which provides wind and solar energy generation time series for past and future scenarios.
However, different GCM-RCM pairings can yield divergent local trends even under the same scenario \citep{Kjellstrom2018}, particularly in complex terrain and coastal regions \citep{Moemken2018}.
Moreover, the PECD remains constrained to selected GCM-RCM pairings, is largely deterministic, and often provides only monthly or aggregated fields rather than the sub-daily spatiotemporal data.

Statistical downscaling, on the other hand, builds empirical relationships between large-scale predictors and local weather and is computationally efficient, but traditional methods often fail to capture the complex spatio-temporal dependencies of multivariate weather fields \citep{Maraun2018,Vandal2019}.
To address this, we adapt a generative deep-learning framework \citep{Schmidt2025} based on score-based diffusion models trained on high-resolution ERA5 to learn joint local weather statistics and to conditionally map coarse CMIP6 fields to spatio-temporally consistent, turbine-relevant realizations.
Crucially, the model's probabilistic sampling produces many plausible high-resolution futures, enabling uncertainty-aware projections of Dunkelflaute occurrence, duration and regional variability across the CMIP6 ensemble.

\section{Data and Methods} \label{sec:methods}

\subsection{Data}
We use the ERA5 reanalysis dataset \citep{ERA5} from the Copernicus Climate Change Service (C3S) for the years 1979 to 2024 with a spatial resolution of 0.25\textdegree and a six hourly temporal resolution.
Six-hourly resolution was shown to be a good compromise between temporal resolution and computational cost \citep{Effenberger2023}.
The variables needed to compute the wind and solar power generation are given in \Cref{tab:era5_cmip6_variables}.
For future projections, we use a selection of the Coupled Model Intercomparison Project Phase 6 (CMIP6) dataset \citep{Eyring2016}: MPI-ESM1-2-HR, GFDL-ESM4, IPSL-CM6A-LR, MIROC6 and CANESM5. The selection is based on their spread in different climate sensitivities \citep{Meehl2020} and take the variables that correspond to ERA5 (see \Cref{tab:era5_cmip6_variables}).
We look at two specific emission scenarios, the Shared Socioeconomic Pathways (SSPs) SSP2-4.5 (referred to as ssp245) and SSP5-8.5 (ssp585), which represent an optimistic low and a high emission scenario, respectively.
The CMIP6 data is downloaded using the Pangeo catalogue\footnote{\url{https://pangeo-data.github.io/pangeo-cmip6-cloud/pangeo_catalog.html}}.
A quantile mapping bias correction is applied to the CMIP6 data using coarsened ERA5 reanalysis as reference.

%TC:ignore
\begin{table}[H]
  \centering
  \caption{Comparison of variable names in ERA5 and CMIP6 datasets.}
  \begin{tabular}{lll}
    \toprule
    \textbf{Variable Name}            & \textbf{ERA5 Name}                   & \textbf{CMIP6 Name / Abbr.} \\
    \midrule
    10m u-component of wind           & 10m\_u\_component\_of\_wind          & uas                         \\
    10m v-component of wind           & 10m\_v\_component\_of\_wind          & vas                         \\
    Surface air temperature           & 2m\_temperature                      & tas                         \\
    Surface solar radiation downwards & surface\_solar\_radiation\_downwards & rsds                        \\
    \bottomrule
  \end{tabular}
  \label{tab:era5_cmip6_variables}
\end{table}
%TC:endignore

\subsection{Estimation of Dunkelflaute Events}
%TC:ignore
\begin{figure}[!htb]
  \centering
  \includegraphics[width=.9\linewidth]{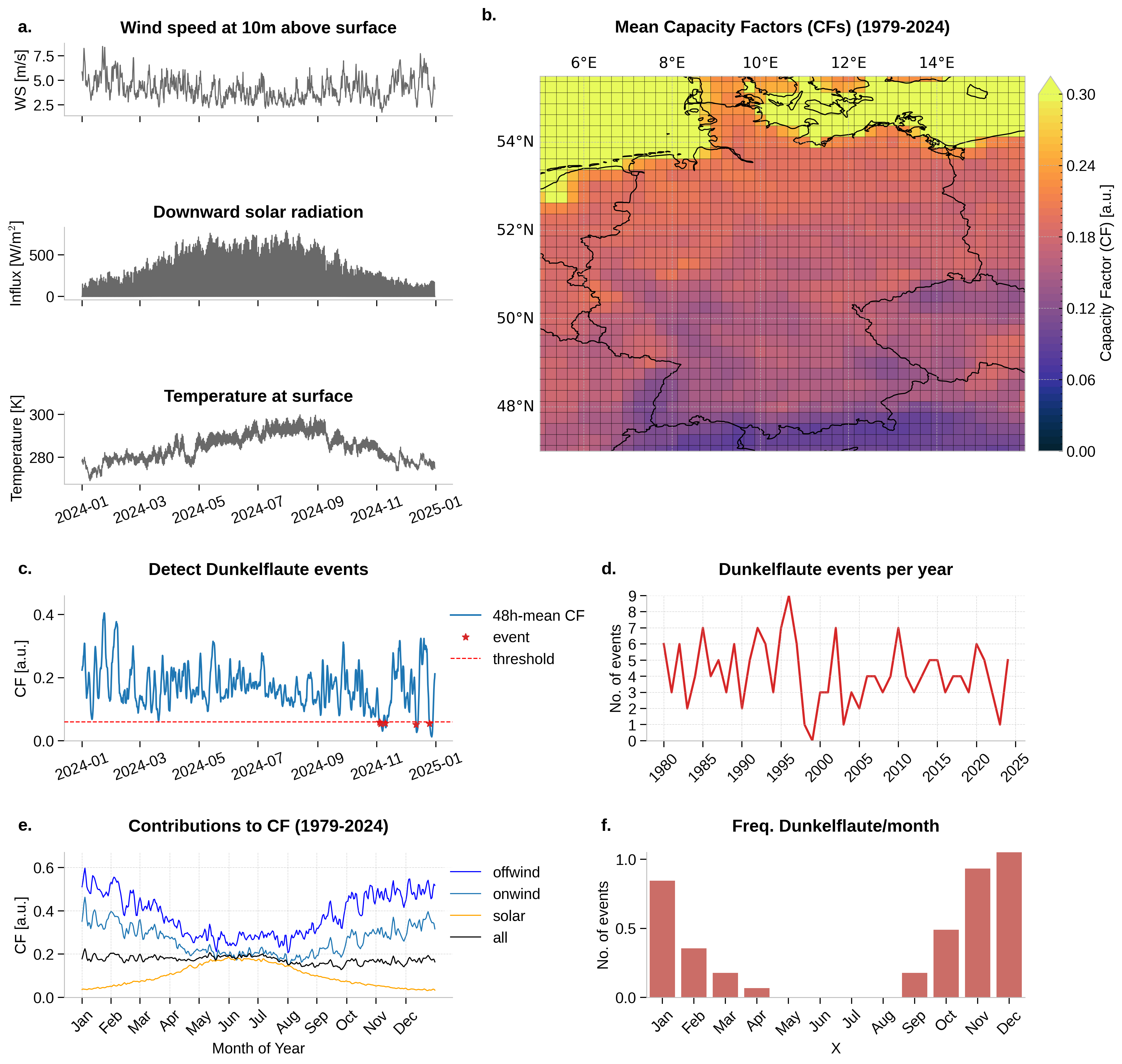}
  \caption{\textbf{Overview of the identification process for Dunkelflaute events in Germany.}
    Panel \textbf{a} displays the time series used to calculate local \glspl{cf} for wind and solar energy in 2024, representative of the approach applied to the full reanalysis dataset. (\textbf{b}) presents the resulting local \glspl{cf} for wind and solar energy across Germany, where the different sources (onshore, offshore, PV) are weighted by their relative fraction in 2024. (\textbf{c}) highlights periods classified as Dunkelflaute events, marked by low wind and solar generation; the red line denotes the threshold of 0.06 used for event detection \citep{Mockert2023}. (\textbf{d}) illustrates the timing and duration of significant low renewable generation periods identified in the ERA5 historical record.
    (\textbf{e}) shows the individual and weighted average \glspl{cf} for onshore, offshore wind and solar energy, considering only grid cells within German borders (with partial cells weighted by their area within Germany, see \Cref{fig:cells_weight_germany}).
    (\textbf{f}) summarizes the monthly frequency of Dunkelflaute events averaged of the period 1979-2024.}
  \label{fig:Dunkelflaute-overview}
\end{figure}
%TC:endignore

Dunkelflaute events are generally characterized as periods when the combined wind and solar power generation falls below a specified threshold of installed capacity for an extended period \citep{Kittel2024}. For event identification, we adopt the methodology described by \citet{Mockert2023}, which relies on analyzing so-called capacity factor time series.

\paragraph{Capacity factor time series}
The capacity factor (\gls{cf}) is defined as the ratio of actual power output $P_{\text{generated}}$ to the theoretical maximum output $P_{\text{max}}$ over a given period, i.e., $CF = \frac{P_{\text{generated}}}{P_{\text{max}}}$. Here, $P_{\text{max}}$ is determined by the installed capacity and the available wind speed or solar irradiance at each time step. A \gls{cf} of $1.0$ indicates operation at full rated capacity for the entire period, while $CF < 1$ reflects periods of reduced generation due to meteorological conditions, outages or maintenance.
Pointwise and spatially averaged \glspl{cf} are calculated using a modified version of the atlite Python package \citep{atlite}, requiring only the variables listed above (\Cref{fig:Dunkelflaute-overview}~a, \Cref{tab:era5_cmip6_variables}).
Since, in this work the \gls{cf} is computed from reanalysis or model data, it only reflects meteorological variability, excluding effects from maintenance or outages such that, for example, $CF = 0.3$ means average output is 30\% of capacity, not constant operation at that level. Thus, \glspl{cf} serves as a robust metric for characterizing renewable intermittency.
Local \gls{cf} time series for onshore wind, offshore wind, and solar are computed per grid cell using a modified atlite package. The total renewable \gls{cf} is then calculated as a weighted average of these, using current installed capacity shares, and this approach is applied consistently for both historical and future periods.
For Germany in 2024, the installed capacities are: 57.7\% solar (99.3 GW), 36.9\% onshore wind (63.5 GW), and 5.4\% offshore wind (9.2 GW) \citep{BNetzA2025Capacities2024}.
The local \glspl{cf} are then averaged over Germany, considering only grid cells within German borders (weighted by their area within Germany, compare \Cref{fig:cells_weight_germany}).
We adopt the atlite package's assumption, which states that the spatial allocation of installed wind and solar capacities is proportional to the respective local capacity factors of each technology.

\paragraph{Dunkelflaute event detection}
Following \citet{Mockert2023}, we define a Dunkelflaute event as a period during which the 48-hour-mean combined \gls{cf} remains below a threshold of $0.06$, shown exemplarily in \Cref{fig:Dunkelflaute-overview}~c for the year 2024 with the above-mentioned events in November and December 2024 (red stars).
Consecutive periods are merged into a single event, placed on the first date of occurrence.
This procedure enables systematic identification of events/year (\Cref{fig:Dunkelflaute-overview}~d). The notable 1997 peak aligns with previous studies \citep{Mockert2023, Kittel2024b}. The same methodology is subsequently applied to the downscaled CMIP6 data (see \Cref{sec:downscaling}).

\paragraph{Dunkelflaute event statistics}
The individual \glspl{cf} for wind and solar energy have a clear seasonal cycle. While off- and onshore winds are on average higher in winter and lower in summer, it is vice versa for solar power (\Cref{fig:Dunkelflaute-overview}~e).
Even though this counterbalance means that on average the capacity remains approximately constant throughout the year (black line in \Cref{fig:Dunkelflaute-overview}~e), the lion's share of Dunkelflaute events occur in winter from November to January (\Cref{fig:Dunkelflaute-overview}~f).
This is likely because solar CFs are substantially lower in winter due to shorter daylight hours and lower irradiance and cannot compensate for days with low wind speeds that can occur in winter as well as in summer.
Consequently, while the average wind \glspl{cf} may be higher in winter than in summer, the combined distribution of wind and solar \glspl{cf} exhibits greater variability and deeper low tails in winter \citep{Li2021, Mockert2023}.

\subsection{Generative Downscaling of CMIP6 Data} \label{sec:downscaling}
\begin{figure}[!htb]
  \centering
  \includegraphics[width=\linewidth]{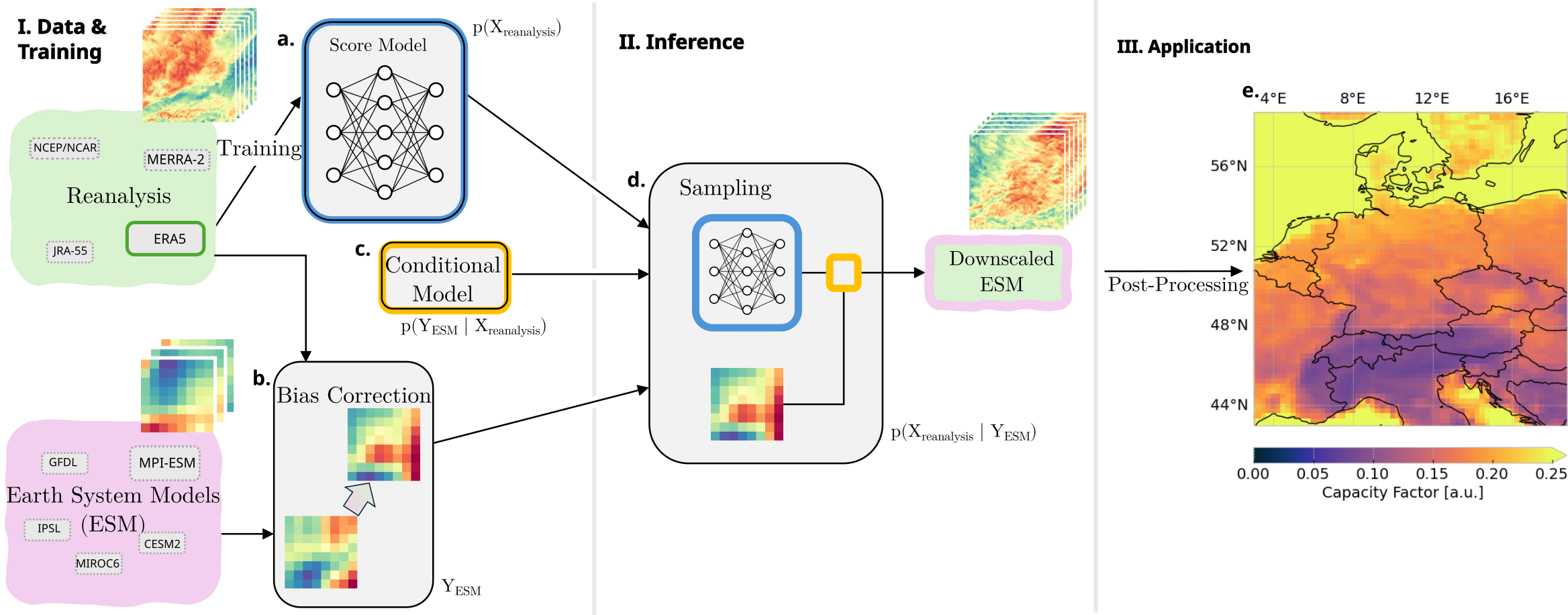}
  \caption{\textbf{Probabilistic pipeline for spatiotemporal downscaling of multiple variables.}
    This schematic illustrates the generative spatio-temporal downscaling framework adapted from \citet{Schmidt2025}. For clarity, only one representative variable is depicted. \textbf{a}: A score-based model is trained on sequences of reanalysis data, enabling the diffusion model to learn fine-scale spatial and temporal patterns. Note that Earth System Model (ESM) simulations are excluded from the training phase. \textbf{b}: An ensemble run from any ESM (here CMIP6, such as MPI-ESM) is selected and pre-processed for downscaling, including an optional bias-correction step to align climate output distributions with reanalysis data. \textbf{c}: The conditional model defines the mapping from the coarse climate simulation ($Y_\mathrm{ESM}$) to the fine-scale reanalysis space ($X_\mathrm{reanalysis}$), establishing constraints for the generative process. \textbf{d}: The trained score model (\textbf{a}) is conditioned on the processed climate input (\textbf{b}) via the observation model (\textbf{c}), generating time series that preserve the statistics and relationships of the coarse climate data.
    \textbf{e}: The generative model samples from the learned distribution, producing fine-scale downscaled time series. These can be further post-processed to obtain \glspl{cf} for wind and solar energy.}
  \label{fig:pipeline-sketch}
\end{figure}

Following recent methodological developments in generative modeling, we train a probabilistic model for time series of atmospheric dynamics.
This work directly builds on the generative spatiotemporal downscaling framework proposed by \citet{Schmidt2025}, which we will abbreviate as GSD (Generative Spatiotemporal Downscaling) in the following.

Panels I and II in \Cref{fig:pipeline-sketch} show a sketch of this framework.
GSD is built around a generative diffusion model (DM) \citep{dm2015sohldickstein,graddata2019song,ddpm2020ho,ddim2021song,sdedm2021song,variationaldms2021kingma,edm2022karras}.
Based on a finite data set, which is assumed to consist of independent draws from a shared probability distribution, DMs are trained to generate new, unseen data points by learning an implicit representation of the unknown data distribution and sampling from it.
In GSD, a DM is trained on sequences of high-resolution reanalysis data (here: ERA5) and acts as a probabilistic model for the output space. During inference (i.e. \textit{after} training), a conditioning mechanism allows for guiding the generated predictions to adhere to flexible boundary conditions provided by coarse climate simulations.
This probabilistic formulation ensures that uncertainty associated with downscaling is reflected through sampling multiple plausible predictions.
\Citet{Schmidt2025} evaluate the model in an on-model setting (i.e. with available ground truth data) and, in a next step, apply it to downscaling output from different members of CMIP6. This work follows a similar approach.

We introduce a few modifications to GSD that preserve the core functionality of the original framework but optimizes the training objective in a simpler and more robust way.
DMs are special instances of the family of flow-based generative models trained with the \emph{Flow Matching} (FM) objective (see \Cref{si:diffusion_fm} for details) \citep{lipman2023flow,lipman2024flowmatchingguidecode}.
While the generative process in the DM formulation follows a stochastic diffusion path guided by the score function to reach regions of high data density, the FM formulation allows for the construction of more general probability paths.
In particular, linear probability paths have been shown to result in more stable training and sampling \citep{lipman2023flow}, which is why this work follows the FM formulation rather than the DM formulation.

GDS relies on training a fully unconditional generative model such that all guiding constraints are imposed \emph{post-training}.
While this allows a high degree of flexibility for downstream tasks, we found it advantageous to provide some conditioning information directly during the training.
Specifically, we train the generative model on pairs of (i) sequences of reanalysis data and (ii) the corresponding time stamps.
The time stamps only include the hour (here: 0:00, 06:00, 12:00, 18:00) and the month (1 through 12).
The reasoning is that the hour of the day and the month capture information about the diurnal and the seasonal cycle.
We train the conditional model by incorporating a learned embedding of the time stamps into the generative model \citep{guidance2021dhariwal}.
It is safe to assume that learning the connection between the dynamics and this temporal information directly results in more informed predictions.
In particular, the hour-of-day information allows for generating temporally-aligned predictions even if the conditioning information aggregates samples longer than 24 hours.

After sampling, we can use the output for any application task (Panel III in \Cref{fig:pipeline-sketch}). Here, we first perform a bias correction on the downscaled data using ERA5 as the reference to compensate for systematic biases introduced by temporal averaging in the input wind data (compare \Cref{fig:distribution_comparison}), which can underestimate sub-daily variability due to daily averaging \citep{Effenberger2023} (see \Cref{fig:distribution_comparison}~a,b).
Further details are provided in \Cref{si:downscaling_bias_correction}.
Next we use the downscaled data to compute generation estimates for wind and solar energy (\Cref{fig:pipeline-sketch}~e).

\section{Results}

\subsection{Evaluation of the downscaling Framework}

\paragraph{Downscaling performance with respect to the ground truth}

\begin{figure}[!htb]
  \centering
  \includegraphics[width=.87\linewidth]{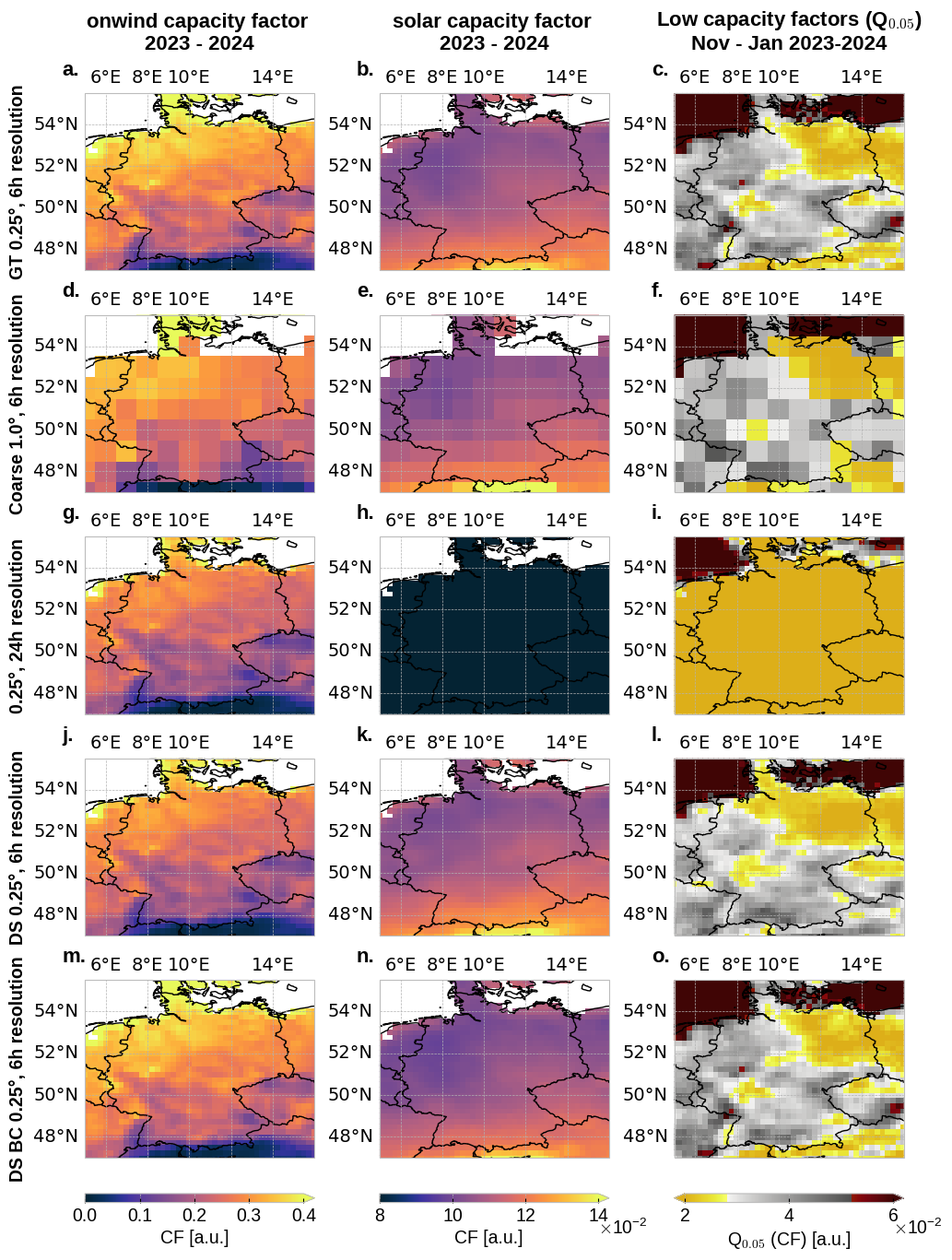}
  \caption{\textbf{Comparison of different spatial and temporal resolutions and the effect of bias correction.}
    The first column presents onshore wind energy \glspl{cf}, the second column shows solar energy \glspl{cf}, and the third column depicts the $0.05$ quantile of the 48-hour running average of \glspl{cf} for November--January.
    Across rows (from top to bottom): \glspl{cf} from ground truth (GT) ERA5 reanalysis data; a spatially coarse (1.0\textdegree) and a temporally coarse (24\,h) resolution of the GT respectively; the uncorrected downscaling (DS) output; the bias-corrected downscaling (DS BC) results.
    The downscaling outputs are averaged over 10 samples drawn from the generative model.}
  \label{fig:compare_res_ds}
\end{figure}

We assess the downscaling framework's ability to accurately reproduce the fine-scale spatial and temporal variability of wind (\Cref{fig:compare_res_ds}~1st column) and solar (\Cref{fig:compare_res_ds}~2nd column) energy capacity factors in Germany.
We further evaluate the model's performance during the critical winter months of November through January (\Cref{fig:compare_res_ds}~3rd column), which are most relevant for Dunkelflaute events.
We find that while spatial downscaling is essential to reproduce spatial variability of wind energy (compare \Cref{fig:compare_res_ds}~d vs \Cref{fig:compare_res_ds}~j,m), temporal downscaling is essential for resolving correct solar capacity factors (\Cref{fig:compare_res_ds}~h) due to the diurnal cycle.
Without these steps, the coarse input data fails to capture the spatial heterogeneity for the critical winter months \Cref{fig:compare_res_ds}~f) or is even unable to obtain a reasonable number of events \Cref{fig:compare_res_ds}~i).
We also see that the bias correction step is essential to avoid an underestimation of the \glspl{cf} for specific regions (e.g. compare North-East Germany in \Cref{fig:compare_res_ds}l vs \Cref{fig:compare_res_ds}c and \Cref{fig:compare_res_ds}o).
The importance of the bias corrected downscaling can also be shown exemplarily for a selected time period of the aforementioned Dunkelflaute events in November 2024 (\Cref{fig:ts_downscaling_comparison}).
While the general trend is captured also by a coarser spatial and temporal resolution, the bias corrected downscaling is closest to the ground truth. The other methods tend to underestimate the \gls{cf} and are therefore likely to overestimate the number of Dunkelflaute events.

\paragraph{Evaluation of the historical CMIP6 downscaled data}
\begin{figure}[!htb]
  \centering
  \includegraphics[width=\linewidth]{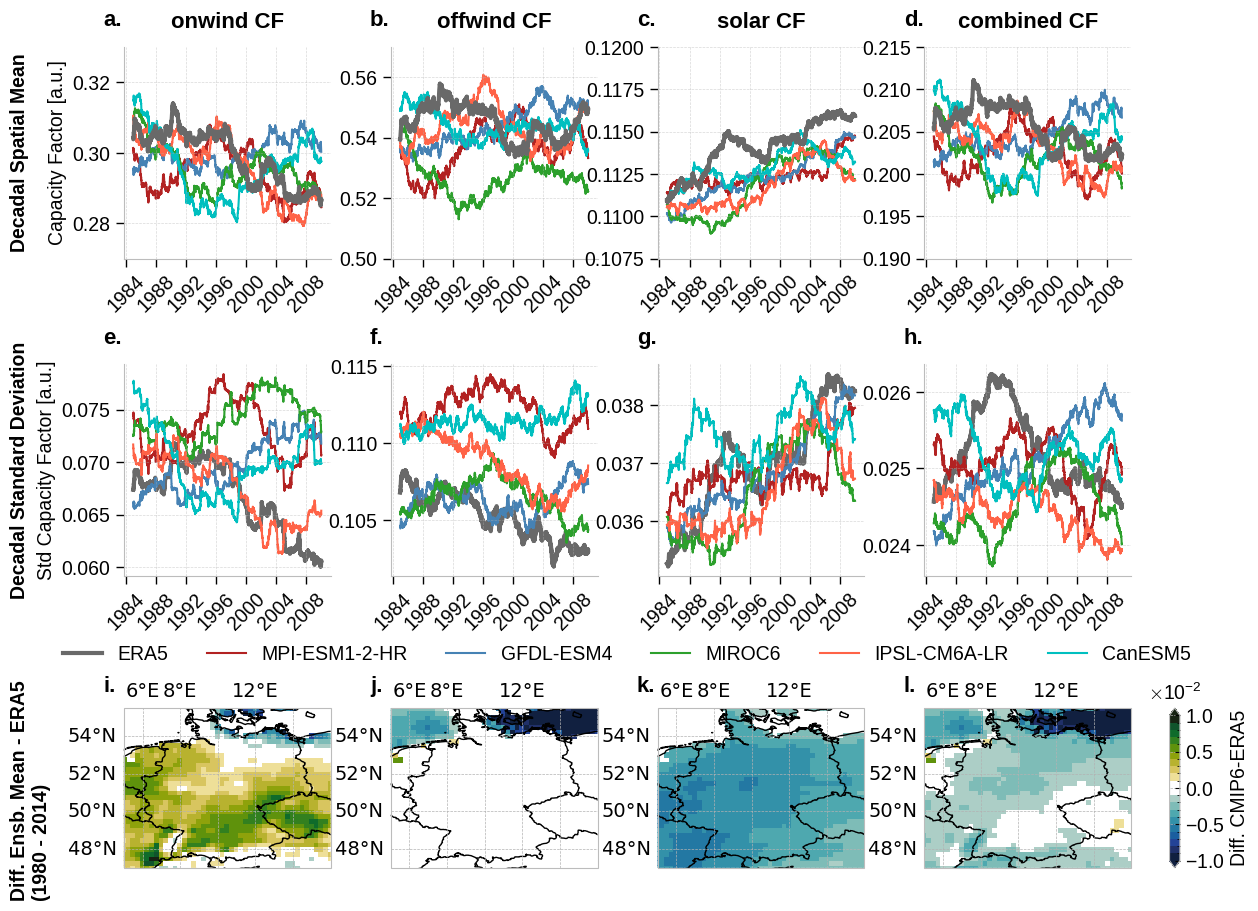}
  \caption{\textbf{Comparison of historical \glspl{cf}s obtained from CMIP6 to ERA5.}
    The figure presents a detailed comparison between the 10-year running average capacity factors (\glspl{cf}) derived from the bias-corrected and downscaled ensemble mean of historical CMIP6 runs and those obtained from ERA5 reanalysis data for the period 1980-2014.
    In the first row, the spatially weighted \glspl{cf} time series for Germany are shown, with individual CMIP6 models represented by colored lines and ERA5 reanalysis by grey lines, allowing for a direct temporal comparison across models and the observational reference.
    The second row denotes the respective decadal standard deviation per time point.
    The third row displays the spatially resolved, temporally averaged differences between the downscaled CMIP6 ensemble mean \glspl{cf}s and ERA5 \glspl{cf}s.
    Specifically, the first column corresponds to onshore wind energy, the second to offshore wind (restricted to grid cells with at least 50\% water coverage), and the third to solar energy (restricted to land grid cells).
    The fourth column shows the weighted mean \glspl{cf}s, where individual sources are combined according to their current relative share in the German energy mix in 2024. The evaluation period covers January 1, 1980, to December 31, 2014.}
  \label{fig:cf_germany_historical}
\end{figure}

We compare the spatial and temporal average \glspl{cf} for wind and solar energy in Germany for the downscaled data of the historical CMIP6 runs to the \glspl{cf}s obtained from ERA5 reanalysis data in the time range 1980-2014, shown in \Cref{fig:cf_germany_historical}.
Since a direct 1:1 comparison between ERA5 and CMIP6 historical model runs is not feasible (ERA5 is an observation-constrained reanalysis that reconstructs the actual historical weather, while CMIP6 historical runs are free-running climate model simulations), we use a 10-year running average to smooth the time series and assess the reproduction of potential temporal trends and variability.
The results show that the time series from ERA5 and the downscaled CMIP6 data show comparable decadal trends and variability.
The 10 year average of onwind \glspl{cf} remained stable around the period, centered from 1985-1998, it decreased in the period since then (1998-2010) by 2 percentage points (\Cref{fig:cf_germany_historical}~a), further the decadal standard deviation is decreasing (\Cref{fig:cf_germany_historical}~e). For the offshore wind \glspl{cf}, the decadal variablity is also decreasing (\Cref{fig:cf_germany_historical}~f), though the \glspl{cf} itself has no clearly visible trend (\Cref{fig:cf_germany_historical}~b).
Contrary to the onwind \glspl{cf}, the solar \glspl{cf} is increasing by 0.5 percentage points throughout the 30 years (\Cref{fig:cf_germany_historical}~c) with an increasing variability.
Even though the weighting of solar \glspl{cf} in the combined \glspl{cf} is largest, the effect of the decrease in onshore wind \glspl{cf} dominates the combined \glspl{cf} and over the 30 year period, a decrease from 1998-2010 by 0.5 percentage points is visible.
The downscaled historical CMIP6 runs reproduce decadal trends in solar \glspl{cf} and show decadal spatial means for onshore and offshore wind \glspl{cf} that remain within the ERA5 range. The decadal standard deviations of wind (\Cref{fig:cf_germany_historical}~e, f), solar (\Cref{fig:cf_germany_historical}~g), and combined (\Cref{fig:cf_germany_historical}~h) \glspl{cf} are also consistent with those derived from ERA5.

A similar pattern is observed on a pixel-wise level.
For onshore winds, the pixelwise deviations (most pixels show only a very slight overestimation of less than 1\%) are marginal for the German mainland (\Cref{fig:local_germany_df}~i).
The coastal region and the Baltic Sea, present an exception from this with a more pronounced deviation between the downscaled and ERA5 \glspl{cf} (\Cref{fig:local_germany_df}~j). These biases are likely inherited from the coarse input data taken from the CMIP6 ensemble, which are known to suffer from inaccuracies at the land-sea transition \citep{Cavaleri2024, Allouche2023}.
We find a slight underestimation for solar energy (\Cref{fig:local_germany_df}~k) that is almost uniformly distributed over Germany, indicating a consistent climate model bias.
Taken together, the results indicate a good agreement for the weighted mean of the different energy sources according to their relative share in the German renewable energy mix (\Cref{fig:local_germany_df}~l).

\subsection{Historical and Future Projections of Dunkelflaute Events}
\begin{figure}[!htb]
  \centering
  \includegraphics[width=\linewidth]{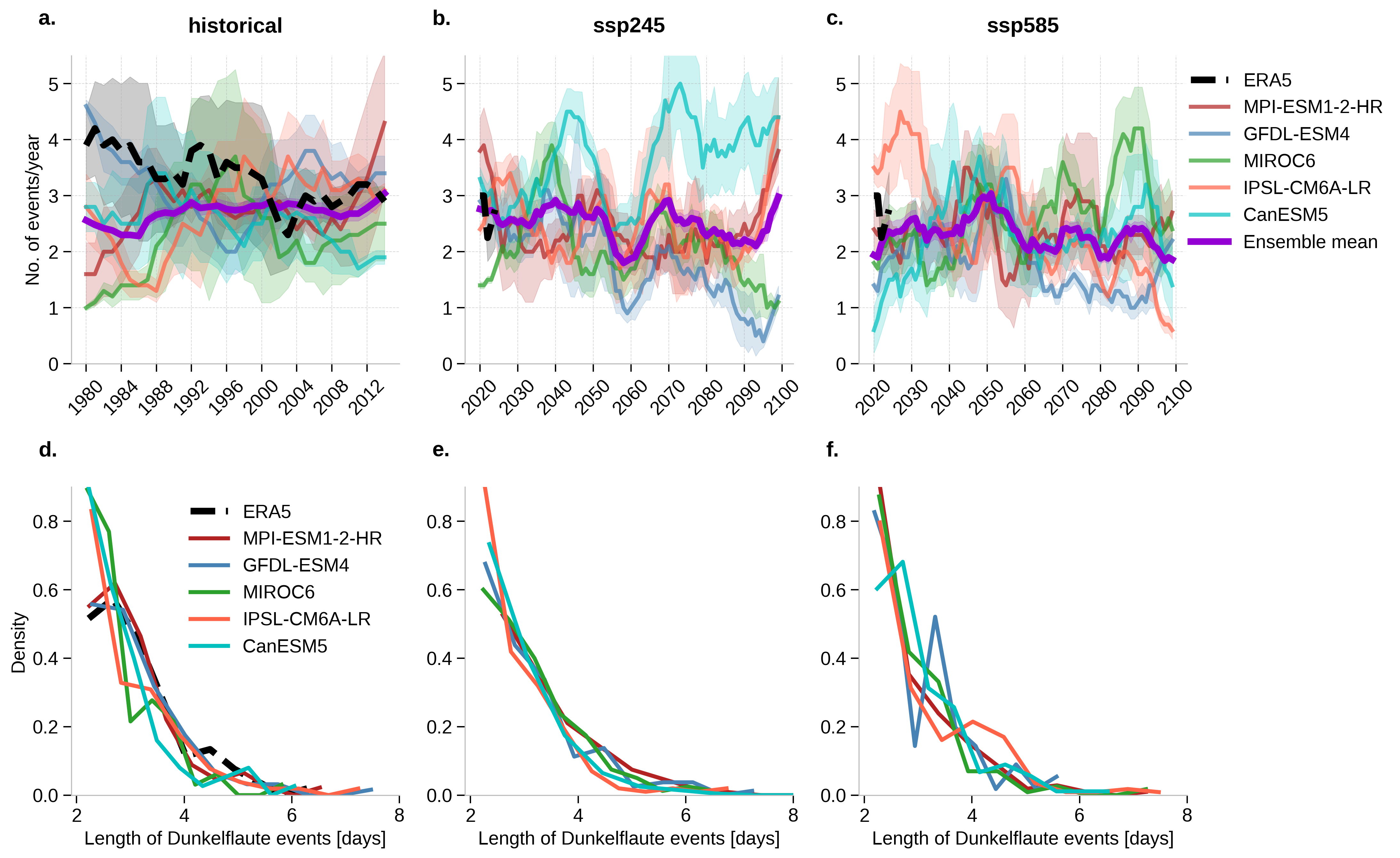}
  \caption{\textbf{Yearly occurrences of Dunkelflaute events in Germany for past and future scenarios.}
    The figure shows (\textbf{a}) a comparison of the historical record to ERA5 and the estimated occurrences of Dunkelflaute events for the time until 2100 for the (b) optimistic-case ssp245 and (c) the worst-case ssp585 emission scenarios.
    The number of events per year is the rolling average over a 10-year window to estimate decadal trends.
    The second row shows the distribution of the duration of all measured events in the respective time period for (\textbf{d}) historical record and (\textbf{e}, \textbf{f}) future scenarios. The errorband denotes the decadal standard deviation.}
  \label{fig:ts_germany_df}
\end{figure}

We first evaluate how well the downscaled CMIP6 data can reproduce the historical record (1980-2014) of occurrences of Dunkelflaute events in Germany using the capacity factor time series (\Cref{fig:ts_germany_df}~a) (see \Cref{sec:methods}).
For this purpose, the number of events per year is shown as a rolling average over a 10-year window to estimate decadal trends.
The downscaled CMIP6 data reveals a similar variability and the range of occurrences per year as ERA5 in the historical record. None of the models exhibit a statistically significant trend (using a Student's t-test with a 0.05 significance level), neither does ERA5. The ensemble mean is from 2000 on very well aligned with the ERA5 data.
We observe different behavior of single models compared to ERA5.
While data from CANESM5 has a consistent decrease but with permanently around 1 event per year less than ERA5, MPI-ESM1-2-HR shows an increase over the historical record.
GFDL-ESM4 and MIROC6 show some decadal fluctuations, while IPSL-CM6A-LR is roughly stable.
The historical ERA5 record lies within the decadal variability of the model ensemble, indicating that the simulations provide plausible scenarios for past Dunkelflaute events.
We further look at the distribution of days per event in the historical record (\Cref{fig:ts_germany_df}~d) and find that the downscaled CMIP6 data shows a comparable distribution of event durations as the ERA5 data, with a slight overestimation for the IPSL-CM6A-LR model in the duration length.
We thus argue that our approach provides a reasonable estimate of the frequency of event occurrences while noting remaining caveats from near-surface wind and downscaling biases that motivate ensemble/probabilistic interpretation rather than single-model forecasts.

Next, we analyze projections for 2020-2100.
Under the low-emission scenario (\Cref{fig:ts_germany_df}~b), taking CANESM5 aside, models are roughly in agreement until 2080, when they start to diverge: MPI-ESM1-2-HR, and IPSL-CM6A-LR catch up with CANESM5 and project an increase, while GFDL-ESM4 and MIROC6 show a slight decrease.
In the high-emission scenario (\Cref{fig:ts_germany_df}~c), Ensemble mean trend remains stable with no significant changes to around 2-3 events per year.
The distribution of duration per event (\Cref{fig:ts_germany_df}~e,f) remains largely unchanged compared to the historical record.
% Overall, the aggregated risk of Dunkelflaute events in Germany is projected to stay stable through most of the century and also the duration is not expected to change significantly, with some model uncertainty after 2080.

\paragraph{Local analysis of prolonged periods of low renewable energy generation}
\begin{figure}[!htb]
  \centering
  \includegraphics[width=.94\linewidth]{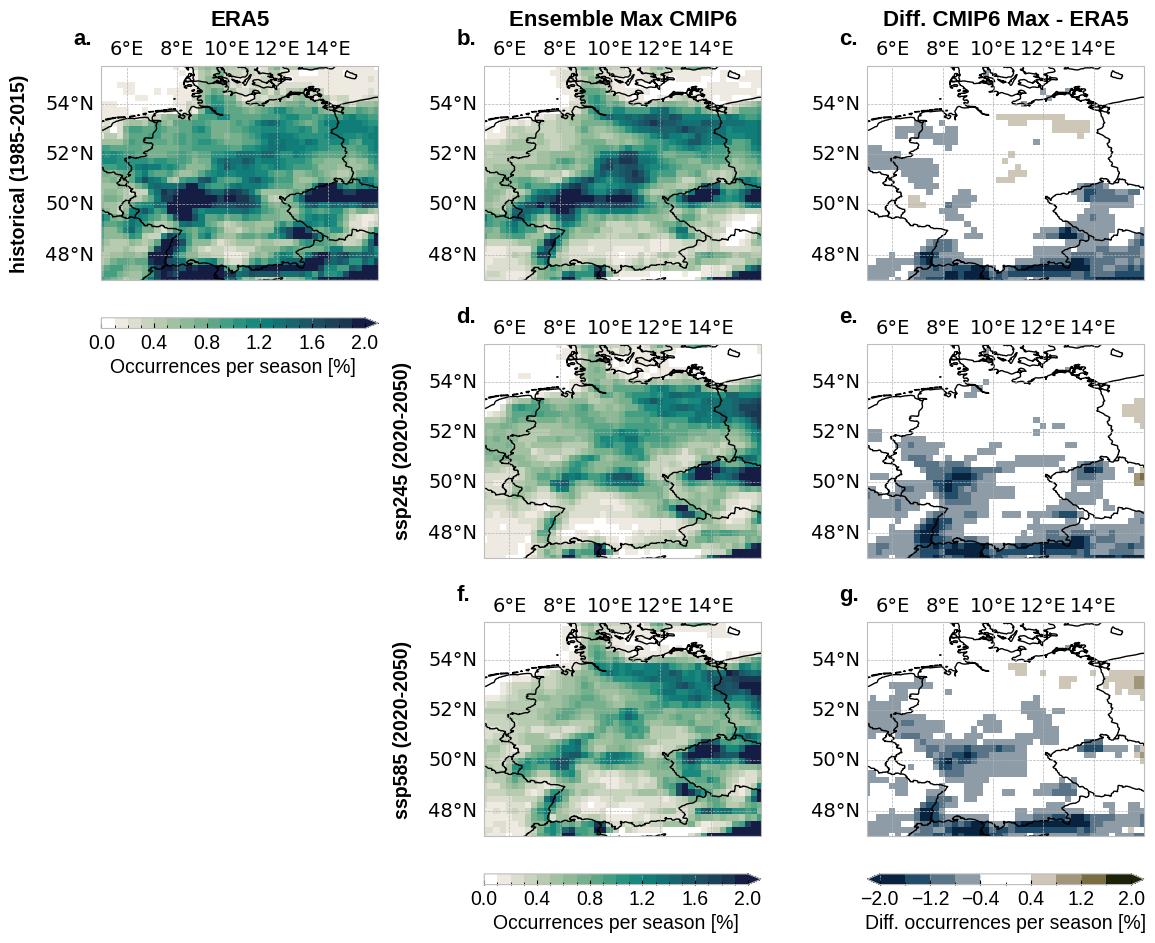}
  \caption{\textbf{Local risk of prolonged (48 hours) low capacity factors (<6\%) for different emission scenarios.}
    The figure shows the spatial distribution of low capacity factors (<6\%) per grid cell for Germany and surrounding regions.
    The first row presents the historical reference for ERA5 (\textbf{a}) and the downscaled CMIP6 ensemble mean for 1985-2014 (\textbf{b,c}). The second and third row display projected occurrences for 2020-2050 under the ssp245 (\textbf{d,e}) and ssp585 (\textbf{f,g}) scenarios, respectively.
    The first column shows ERA5 reference, the second column the CMIP6 ensemble maximum, and the third row the difference to the ERA5 reference period.}
  \label{fig:local_germany_df}
\end{figure}

While the ensemble mean of CMIP6 suggests no significant change in the overall number of Dunkelflaute events for Germany, notable local variations in the frequency of low capacity factor days may still occur.
This spatial analysis is especially important for strategic planning of wind and solar installations, as it identifies regions where the risk of extended low renewable generation may shift over the typical operational lifetime of assets (20-30 years) necessary for informed investment decisions by highlighting areas that may become less optimal under projected future climate conditions.
In \Cref{fig:local_germany_df}, we assess the pixelwise occurrences of prolonged periods ($\geq48$ hours) of low combined capacity factors (<6\%) for the near future (2020-2050).
The historical reference from ERA5 (\Cref{fig:local_germany_df}~a) shows a heterogeneous spatial distribution across Germany, with a general concentration from southwest to northeast and over uplands and mountains.
The CMIP6 ensemble mean for the historical period (\Cref{fig:local_germany_df}~b) broadly replicates this pattern, albeit with problems over mountain regions such as the Alps (\Cref{fig:local_germany_df}~c).
This bias arises as wind speeds suffer from a bias towards lower values in these mountainous regions \citep{Moemken2018, Shen2022, Lafferty2023}, resulting in more periods of low output. However, due to topography, wind turbines are rarely installed there, so this can be neglected for our analysis.
For the future scenarios, analysis of the ensemble mean indicates that the spatial distribution of event frequency per grid cell remains largely unchanged (\Cref{fig:local_germany_df}~d,f) compared to the historical reference, suggesting no substantial alteration in local patterns under future projections.
In the future scenarios, we find a consistent decrease in event frequency in the Hessian region (\Cref{fig:local_germany_df}~e, g) and a slight decrease in the northeast near the Polish border.
Importantly, the differences in event frequency among the ensemble mean, maximum, and minimum are comparable to the inter-model variability (\Cref{fig:local_germany_df_mean}~j,k,l).
These results indicate that the overall frequency of events in Germany is projected to remain in most regions on a similar level, although localized changes may occur, particularly in Southwestern Germany and near the Polish border.

\section{Discussion and Conclusion}
The present study adapts a recently developed generative downscaling framework \citep{Schmidt2025} to assess local changes in the risk of Dunkelflaute events projected by the CMIP6 ensemble.
Our results indicate that there is a notable spread in the CMIP6 model ensemble, while in the ensemble mean the spatially averaged occurrences of periods of combined low wind and solar power generation in Germany is not expected to change substantially in the future.
The pixelwise analysis reveals that certain areas, particularly in the eastern parts of Germany and Poland, may experience a reduction in the wind speeds and might therefore become less attractive for newly installed wind power stations, while part in the Southwestern area of Germany are projected to become more attractive in the future.
This is largely in agreement with other studies.
Even though CMIP6 projections consistently indicate a decline in mean near-surface wind speeds across most of Europe over the 21st century, with reductions of around 5-15\% under SSP2-4.5 and up to 20\% under SSP5-8.5, particularly in northern and central regions \citep{Martinez2024}, for Germany, however, both raw CMIP6 output and downscaled analyzes suggest only minor changes in mean wind resources, with some small increases under SSP2-4.5 \citep{Effenberger2025}.
More critical for the energy system is variability: climate change is expected to make very rare but multi-day long wind droughts by 20-40\% more likely\citep{Qu2025} while intensifying winter storm winds in Northern Europe, both of which pose challenges for grid stability and adequacy \citep{Little2023}.
Such effects could not be identified in this study but were not systematically examined and remain subject to future research.

The downscaling framework presented here is a powerful tool for generating high-resolution climate projections that can be used to assess the potential impacts of climate change on renewable energy resources.
However, this work focuses solely on the weather-induced likelihood of periods of combined low wind and solar power generation, which is only one aspect of the overall risk assessment for renewable energy systems.
Other factors, such as the availability of storage technologies, grid infrastructure, and demand-side management strategies, also play a crucial role in determining the overall resilience of the energy system \citep{Kittel2024}.
Future work is therefore needed to integrate these aspects into a comprehensive risk assessment framework for renewable energy systems.
One potential candidate would be the PyPSA package \citep{pyPSA}. It has already been used to simulate the European energy grid's transition towards renewable energy sources \citep{Horsch2018, Frysztacki2020} and could be used to assess the impact of Dunkelflaute events on the grid's stability.
These should then be coupled with local demand time series and storage capacities. To explore the grid stability, one could consider coupling the downscaled data with a grid simulation model that simulates the flow of electricity through the grid.

The methodology presented here is general and can be applied to any region by training the generative downscaling framework on local high-resolution reanalysis data and conditioning on relevant coarse climate model outputs. This enables assessment of Dunkelflaute risks or similar meteorological events in different geographic contexts.
The presented climate change impact assessment uses the CMIP6 ensemble as input. These are tied to hard-coded emission pathways, the so-called Shared Socioeconomic Pathways (SSPs) and Representative Concentration Pathways (RCPs), i.e., standardized scenarios describing different future trajectories of global societal development (e.g., population growth, economic development, technological progress) combined with different levels of greenhouse gas emissions.
Thus, they  offer only limited flexibility for ``what-if'' questions outside the existing SSP-RCP matrix.
Future work will therefore address this limitation by further developing our framework \citep{Schmidt2025} into a conditional generative one that, for specified conditions (e.g., +1.5~\textdegree C warming), produces physically plausible coherent ensembles of meteorological fields and translates them into impact-ready variables - enabling rapid, flexible, and high-fidelity scenario generation tailored to decision-maker needs.

%TC:ignore

\section*{Data Availability}

Datasets for the observational data from 1979 to date were taken from Copernicus Climate Change Service (C3S) (https://cds.climate.copernicus.eu/cdsapp\#!/dataset/reanalysis-era5-pressure-levels?tab=overview) \citep{ERA5}.
The CMIP6 data were obtained from the Pangeo catalogue (\url{https://pangeo-data.github.io/pangeo-cmip6-cloud/})

\section*{Code Availability}
The code for the downscaling framework is available at https://github.com/schmidtjonathan/DunkelFlowten.
The analysis uses the geoutils package \citep{geoutils}.
The analysis scripts can be found at https://github.com/fstrnad/dunkelflauten.git.

\section*{Acknowledgements}
The project is supported by the German Federal Ministry of Research, Technology and Space (BMFTR) through the FEAT project (grant number 01IS22073B).
The authors gratefully acknowledge financial support by the DFG Cluster of Excellence "Machine Learning - New Perspectives for Science", EXC 2064/1, project number 390727645 and the German Federal Ministry of Education and Research (BMBF) through the Tübingen AI Center (FKZ: 01IS18039A).
F.S. and N.L. acknowledge funding by the Deutsche Forschungsgemeinschaft (DFG, German Research Foundation) under Germany's Excellence Strategy - EXC 2064/1.
J.S. thanks the International Max Planck Research School for Intelligent Systems (IMPRS-IS) for supporting his PhD program.

\section*{Author's contribution}
The joint project was initiated by F.S. and N.L. and coordinated by P.H. and N.L.
J.S. implemented the code base for the  model and training. F.S. and J.S. handled data-processing and -pipeline (reanalysis data) and designed the experiments.
F.S. implemented the data-processing for the climate simulations.
F.S. was responsible for evaluation and visualization of the results.
All authors discussed the results.
The first version of the article was written by F.S., after which all authors reviewed and edited the manuscript.

\section*{Competing Interests}
The authors declare that they have no competing interests.
% \clearpage
% \newpage
\bibliography{./library.bib}% Produces the bibliography via BibTeX.

\newpage
\appendix
\renewcommand{\thefigure}{S\arabic{figure}}
\renewcommand\thesection{SI~\arabic{section}}
\setcounter{figure}{0}
\section*{Supplementary Information}
\section*{Introduction}

This Supplementary Information (SI) provides additional details and supporting analyses for the main manuscript.
We begin by evaluating the performance of the downscaling framework through a comparison of variable distributions between ground truth and downscaled samples (\Cref{si:downscaling_bias_correction}, \Cref{fig:distribution_comparison}), followed by a visual example illustrating the reconstruction of fine-scale spatial temperature patterns for a representative day (\Cref{fig:downscaling-example-tas}).
Next, we describe the methodology for computing national average capacity factors (\Cref{si:capacity_factors}), highlighting the spatial weighting of grid cells within Germany (\Cref{fig:cells_weight_germany}), and present the spatial distribution of wind and solar capacity factors across the country (\Cref{fig:cf_map_germany}). We then compare time series of combined wind and solar capacity factors at different spatial and temporal resolutions, demonstrating the importance of high-resolution data and bias correction (\Cref{fig:ts_downscaling_comparison}).
For future projections, we assess in \Cref{si:future_dunkelflaute} the local risk of prolonged low-capacity-factor events under different emission scenarios, showing both the spatial distribution of event frequency and the inter-model variability across the ensemble (\Cref{fig:local_germany_df_mean}).
Finally, we provide further methodological details on the generative downscaling approach, including the flow matching formulation and time conditioning strategies, to support the reproducibility and transparency of our results (see Section~\ref{si:diffusion_fm}).

\section{Downscaling, distributions and bias correction}  \label{si:downscaling_bias_correction}
\begin{figure}[!htb]
  \centering
  \includegraphics[width=0.8\linewidth]{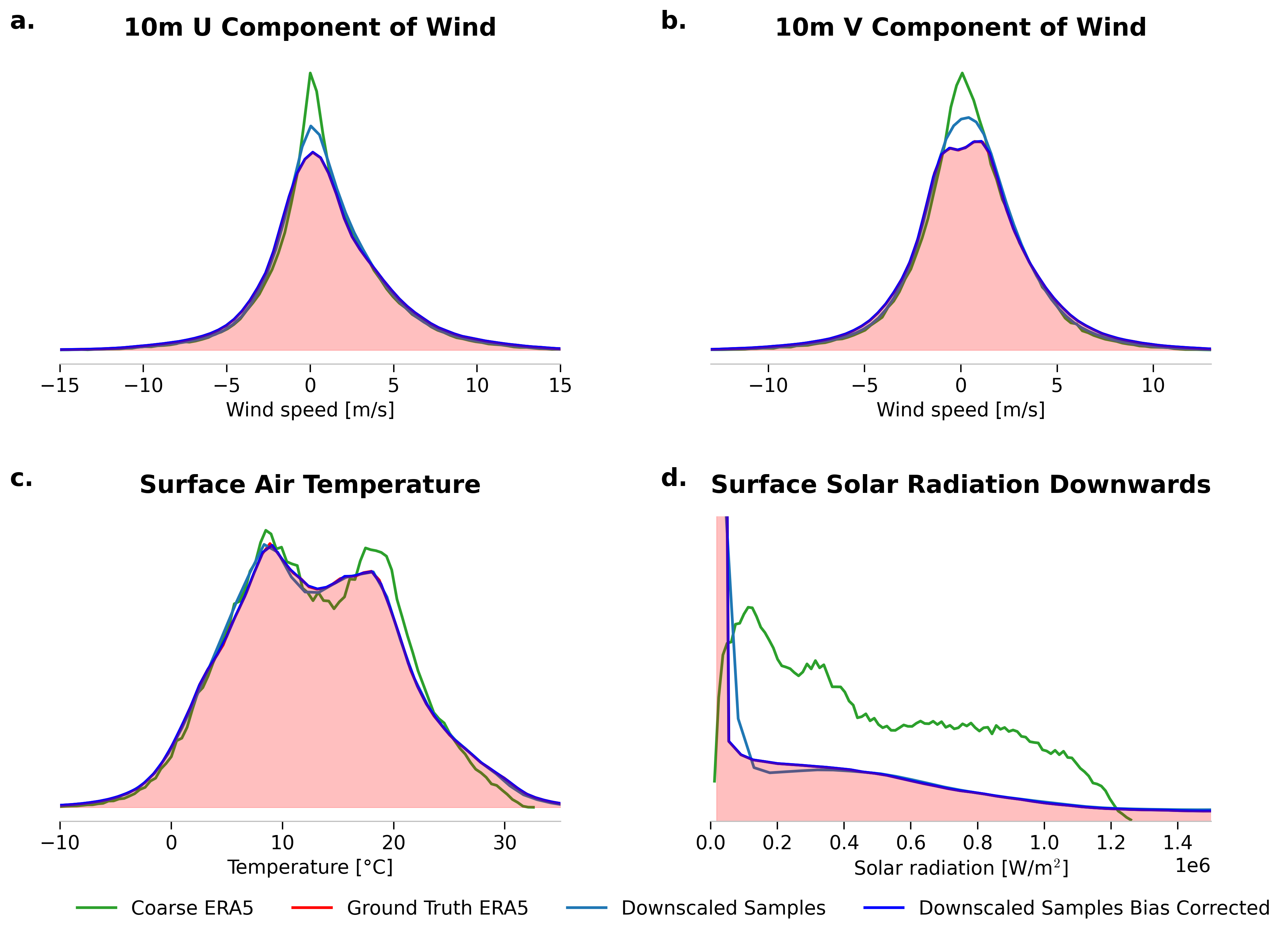}
  \caption{\textbf{Comparison of ground truth and downscaled values for key variables.}
    The figure shows the performance of the downscaling framework for four variables: (a) $u$-component of wind, (b) $v$-component of wind, (c) surface air temperature, and (d) downward solar radiation at the surface. For each variable, the distribution of ground truth (ERA5 reanalysis, red shading) values is compared to the downscaled output (light blue: Samples, bright blue: post-processed bias corrected Samples). The coarse input data (ERA5, artificially coarsened to $1.0\textdegree$, daily resolution, green) is also shown for reference.
    The evaluation period ranges from 1st January 2023 to 31st December 2024.}
  \label{fig:distribution_comparison}
\end{figure}

The downscaling framework is evaluated by comparing the distribution of the samples, obtained from downscaling the artificially coarsened data to the ground truth (ERA5 reanalysis, $0.25\textdegree$ resolution). The results are shown in \Cref{fig:distribution_comparison}.
Overall, the downscaled values match the distribution of the ground truth (red shadding), indicating that the framework captures the fine-scale spatial and temporal patterns of the variables.
Nevertheless, we find that the downscaled values for the $u$-component of wind (\Cref{fig:distribution_comparison}a), and $v$-component of wind (\Cref{fig:distribution_comparison}b) slightly overestimate the distribution around the zero wind speed.
This artifact is a known problem that arises from temporal averaging of wind data \citep{Effenberger2023}, as the input data in the form of daily averages ignores the variability through the day-night cycle (compare the light blue lines to the ground truth red line in \Cref{fig:distribution_comparison}~a,b).
We therefore apply a further post-processing bias correction step to the downscaled values, which leads to a better match with the ground truth (compare bright vs light blue lines to the red line in \Cref{fig:distribution_comparison}~a,b).

For surface air temperature (\Cref{fig:distribution_comparison}c), the downscaled samples (light and bright blue lines) closely match the ground truth, indicating that the framework effectively captures the fine-scale spatial patterns of temperature. The downward solar radiation at the surface (\Cref{fig:distribution_comparison}d) is also well captured by the downscaled values, with a slight overestimation of the distribution around $0.05-0.1$ $10^6$W/m$^2$.
The performance of the downscaling for the downward solar radiation is, nevertheless, surprisingly good, considering that the coarse input data has a very different distribution (compare the light blue lines to the green lines) due to the daily averaging of the input data.

\begin{figure}[!htb]
  \centering
  \includegraphics[width=.8\linewidth]{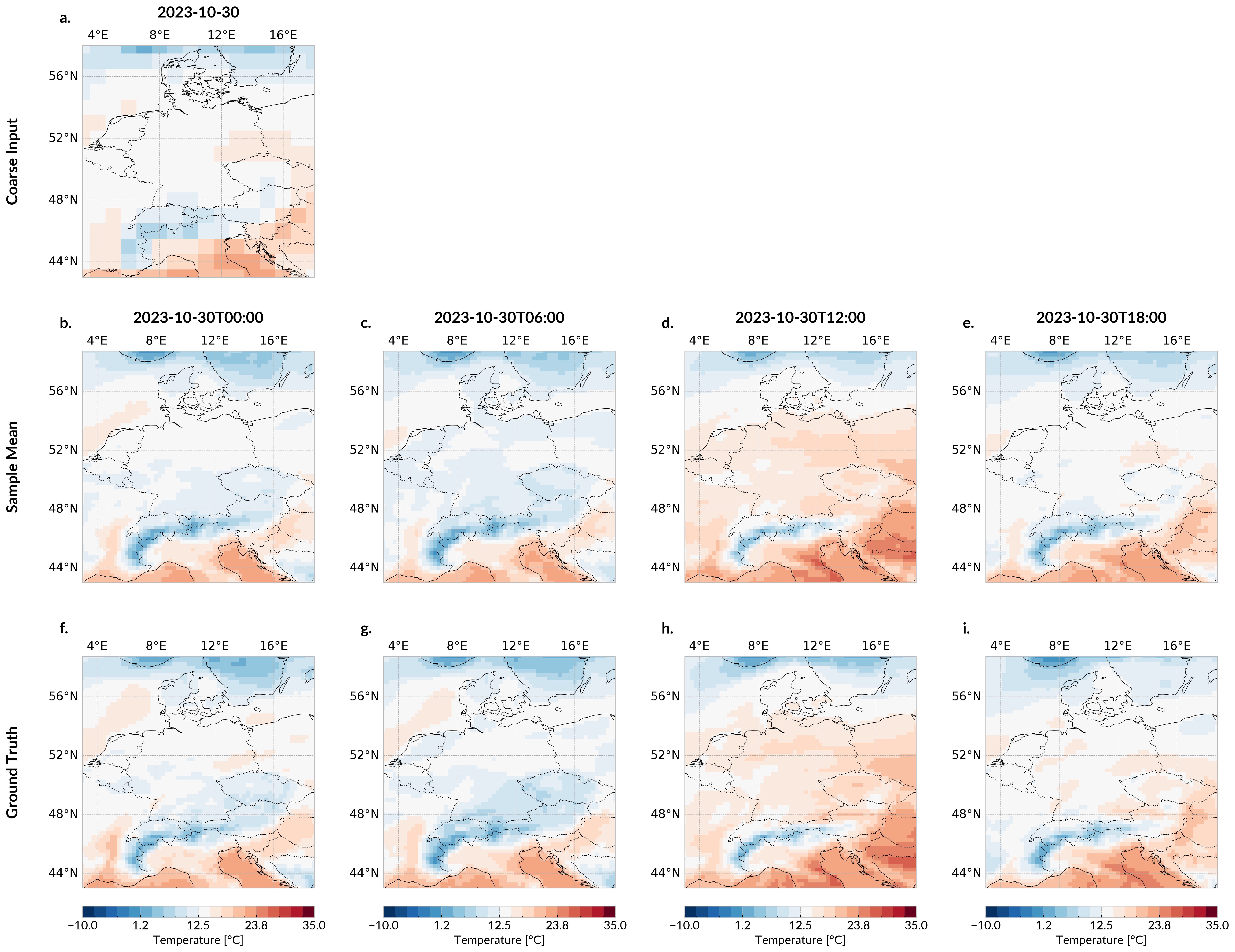}
  \caption{\textbf{Visual example of the downscaling process for surface air temperature on 2023-10-30.}
    The figure compares (a) the coarse-resolution input, (b) the downscaled output, and (c) the ground truth (ERA5 reanalysis) for surface air temperature over Germany on 10th October 2023.
    The downscaling framework successfully reconstructs fine-scale spatial patterns that are absent in the coarse input, closely matching the ground truth.}
  \label{fig:downscaling-example-tas}
\end{figure}

An illustrative example for the spatio-temporal downscaling is presented for a representative day, 10th October 2023 in \Cref{fig:downscaling-example-tas}. The visual comparison confirms that the downscaled fields show a strong agreement with the ground truth, indicating that the framework effectively captures temporal variability, including the diurnal cycle.
Overall, we are confident that the downscaling framework successfully reconstructs the fine-scale spatial patterns.

\section{Capacity Factors} \label{si:capacity_factors}
Capacity factors (CF) are computed per grid cell and per technology (onshore wind, offshore wind, PV) and then aggregated to a national time series. For a single cell i and technology k at time t we define
\[
  CF_{i,k}(t) \;=\; \frac{P_{i,k}(t)}{P^{\max}_{i,k}},
\]
where $P_{i,k}(t)$ is the instantaneous power output estimated from the meteorological fields (wind components, surface temperature, downward shortwave radiation) using the atlite conversion routines (hub-height extrapolation and technology power curves) and $P^{\max}_{i,k}$ is the installed capacity assigned to that cell/technology.

Installed capacity is spatially allocated within Germany proportional to the long-term mean local resource for each technology: the nominal capacity assigned to cell i and technology k is
\[
  C_{i,k} \;=\; C^{\mathrm{tot}}_{k}\,\frac{\overline{CF}_{i,k}}{\sum_{j\in\mathrm{DE}}\overline{CF}_{j,k}},
\]
where $C^{\mathrm{tot}}_{k}$ is the total installed capacity of technology k (national share) and $\overline{CF}_{i,k}$ is the long-term mean CF for that cell. Cells intersecting the national border are included with an area fraction weight $a_i\in(0,1)$.

\begin{figure}[!htb]
  \centering
  \includegraphics[width=.5\linewidth]{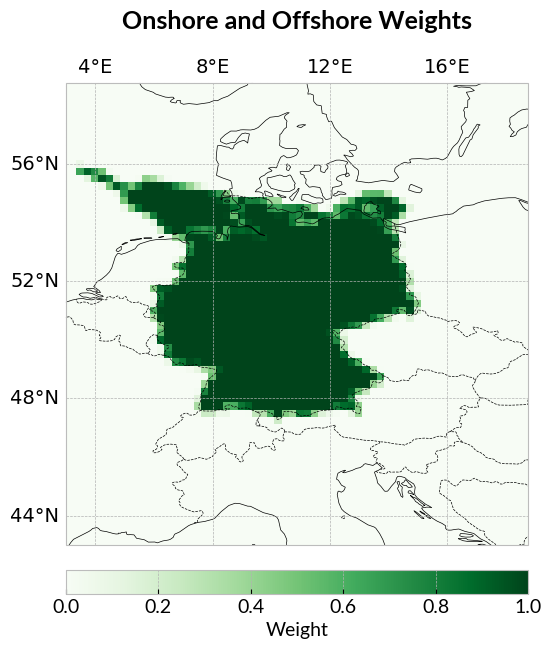}
  \caption{\textbf{Visualization of the contribution of each grid cell to the total \gls{cf} time series in Germany.}
    This figure shows the spatial distribution of the weights assigned to each $0.25\textdegree$ grid cell within Germany when computing the national average \gls{cf} time series. Cells at the border of Germany are only partially included in the average, and their contribution is weighted by the area of the cell that is within Germany.}
  \label{fig:cells_weight_germany}
\end{figure}

The national (area- and capacity-weighted) combined CF time series is then
\[
  CF_{\mathrm{DE}}(t) \;=\; \sum_{k}\sum_{i\in\mathrm{DE}} w_{i,k}\,CF_{i,k}(t),
  \qquad
  w_{i,k} \;=\; \frac{C_{i,k}}{\sum_{k}\sum_{i\in\mathrm{DE}} C_{i,k}},
\]
so that the weights sum to one. The weighting factors are show in \Cref{fig:cells_weight_germany}. The total installed capacities per technology are taken from the German Federal Network Agency (Bundesnetzagentur) for the year 2024: 60.4 GW onshore wind, 7.7 GW offshore wind, and 66.5 GW PV \citep{BNetzA2025Capacities2024}.

\begin{figure}[!htb]
  \centering
  \includegraphics[width=1.\linewidth]{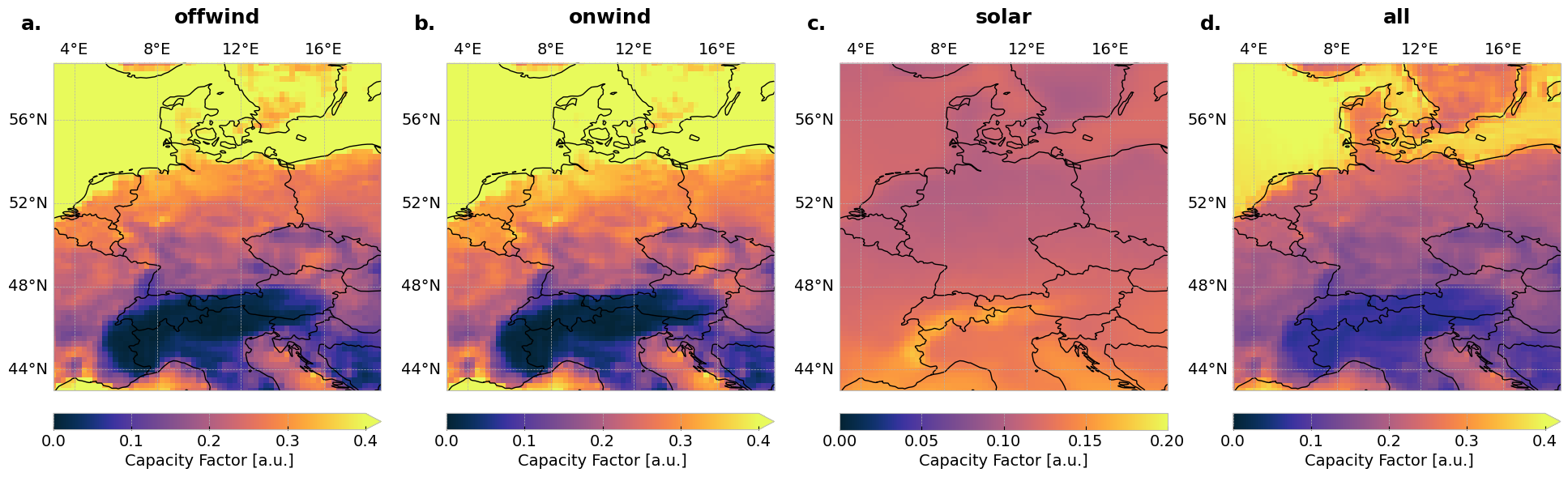}
  \caption{\textbf{Capacity factors for wind and solar energy in Germany.}
    This figure shows the spatial distribution of the capacity factors for wind and solar energy across each $0.25\textdegree$ grid cell within Germany. Panel \textbf{a} shows the offshore wind capacity factors (only over sea areas), \textbf{b} the onshore wind capacity factors, \textbf{c} the solar capacity factors, and \textbf{d} the weighted mean capacity factor, where the offshore wind is only considered over sea grid cells.}
  \label{fig:cf_map_germany}
\end{figure}

The capacity factors for the different technologies are shown in \Cref{fig:cf_map_germany}.

\begin{figure}[!htb]
  \centering
  \includegraphics[width=.8\linewidth]{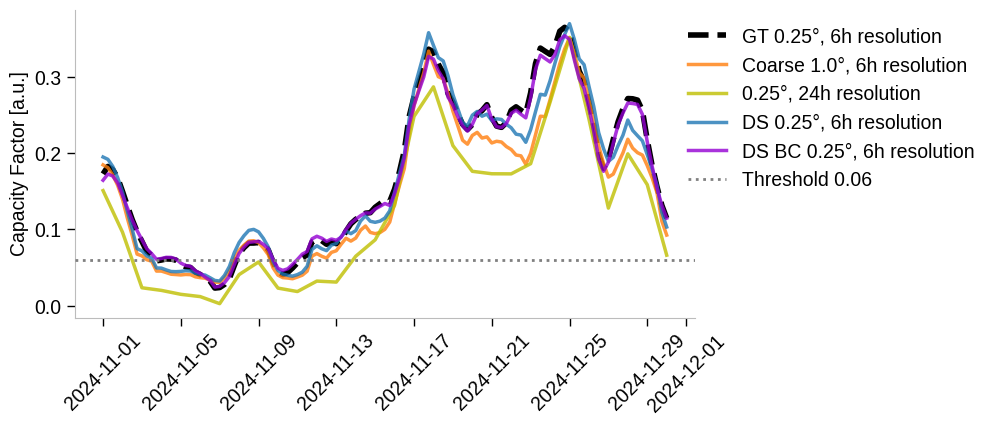}
  \caption{\textbf{Time series of capacity factors for combined wind and solar energy in Germany for different spatial and temporal resolutions.}
    Different resolutions capture varying degrees of temporal variability, with higher resolutions providing more detailed representations of short-term fluctuations. We compare the time series of capacity factors for combined wind and solar energy in Germany for different spatial and temporal resolutions. The evaluation period spans November 1, 2024, to December 1, 2024, using data not included in training.
    The dashed black line represents the ground truth (ERA5 reanalysis data at $0.25$\textdegree spatial and hourly temporal resolution). The orange line shows results for a coarser spatial resolution ($1.0\textdegree$) but the same temporal resolution (6 hourly). The yellow line shows results for the same spatial resolution ($0.25\textdegree$) but a coarser temporal resolution (daily). The light blue line shows results for the downscaled coarse spatial ($1.0\textdegree$) and temporal (daily) resolution. The purple line shows the output from the bias corrected downscaling framework, averaged over 10 samples drawn from the generative model.}
  \label{fig:ts_downscaling_comparison}
\end{figure}

In \Cref{fig:ts_downscaling_comparison}, we compare time series of combined wind and solar capacity factors at different spatial and temporal resolutions, demonstrating the importance of high-resolution data and bias correction.

\section{Future Dunkelflaute events} \label{si:future_dunkelflaute}
\begin{figure}[!htb]
  \centering
  \includegraphics[width=1.\linewidth]{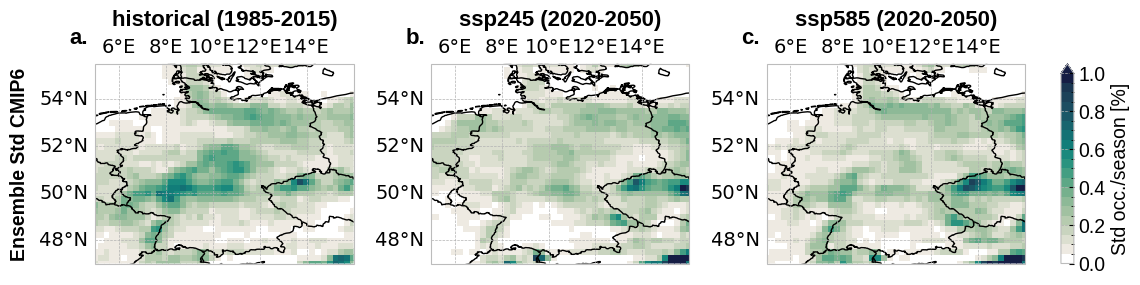}
  \caption{\textbf{Standard Deviation of local risk of prolonged (48 hours) low capacity factors (<6\%) for different emission scenarios.}
    The row illustrates the inter-model variability across the ensemble for the spatial distribution of low capacity factors (<6\%) per grid cell for Germany and surrounding regions. The first column presents the downscaled CMIP6 ensemble mean for 1980-2014. The second and third columns display projected Dunkelflaute occurrences standard deviation under the ssp245 (optimistic emission reduction) and ssp585 (high emission) scenarios, respectively.}
  \label{fig:local_germany_df_mean}
\end{figure}

In \Cref{fig:local_germany_df_mean}, we assess the inter-model variability across the ensemble for the spatial distribution of low capacity factors (<6\%) per grid cell for Germany and surrounding regions.
We find that the standard deviation across the ensemble is of similar magnitude as the differences between the ensemble mean and the ERA5 reference period, indicating that the uncertainty across models is substantial and indicating that the wind fields are highly variable.

\section{Methods} \label{si:diffusion_fm}

\paragraph{Flow matching}

% TODO (J.S.): Write intro to flow matching

The framework by \citet{Schmidt2025} is based on diffusion models \citep{{dm2015sohldickstein,graddata2019song,ddpm2020ho,ddim2021song,sdedm2021song,variationaldms2021kingma,edm2022karras}}, which generate samples by following a stochastic diffusion process $\{\rX(\difftime); \difftime \geq 0\}$ that transports a tractable noise distribution $p_T(\rX(T))$ (e.g.\,Gaussian) to a complex data distribution $p(\rX(0))$.
In our statistical downscaling setting, the target distribution is that of sequences of high-resolution reanalysis data.
The dynamics of the generative diffusion process are governed by a \emph{score function} $s(\rX(\difftime), \difftime) \approx \nabla_{\rX(\difftime)}\log p_\difftime(\rX(\difftime))$.
The score function is approximated by a learned parametric model (neural network) $s^{(\theta)}$, such that the parameters $\theta$ minimize a de-noising score matching objective \citep{denoisingsm2011vincent,ddpm2020ho}.
\Citet{Schmidt2025} used a subset of the COSMO-REA6 \citep{bollmeyer2015towards} reanalysis product as training data.
Sampling posterior and sequential estimates require extending the standard DM framework, accordingly.
For this purpose, we follow the work of \citet{Schmidt2025} and build upon the score-based data assimilation (SDA) framework proposed by \citet{sda2023rozet}.
SDA provides a framework for sampling from an approximate posterior over length-$L$ state sequences $(X_1, \dots, X_L) =: X_{1:L}$, i.e.,
\begin{equation}
  p(\rX_{1:L}(0) \mid \rY).
\end{equation}
The conditioning information $\rY$ is related to the inferred sequence via a statistical relationship, or observation model,
\begin{equation}
  p(\rY \mid \rX_{1:L}).
\end{equation}
This setup casts statistical downscaling as a Bayesian inverse problem defined by the observation model and the trained DM, which acts as a prior.

Instead of stochastic processes, \citet{lipman2023flow} show how a specific simple choice of probability path that linearly interpolates between data and noise distributions makes both training and generation more robust.
These results motivate the present work adopting this formulation, in which a vector field $v$ replaces the score model $s$.
Notably, under standard assumptions, the vector field $v$ is related to the score function $s$ via a simple affine transformation \citep[Section 4.10]{lipman2024flowmatchingguidecode}.
Therefore, the posterior-sampling mechanism (cf.~\citep[Eq.\,(5)]{Schmidt2025}) directly transports from the score-based to the flow-based formulation.

\paragraph{Time Conditioning}

In this work, we train a conditional flow model that receives temporal information about the current hour of day $t\hod$ and the current month $t\mnth$.
Concretely, we train a score model
\begin{equation}\label{eq:cond-flow}
  v^{(\theta)}(\rX(\difftime), \difftime, t\hod, t\mnth).
\end{equation}
The generative process that is guided by \cref{eq:cond-flow} approximately samples from a conditional distribution
\begin{equation}\label{eq:time-conditional-score-model}
  p(\rX \mid t\hod, t\mnth).
\end{equation}
The conditional model is trained following the "classifier-free guidance" mechanism proposed by \citet{ho2021classifierfree}.
This requires training the flow model $v^{(\theta)}$ on pairs of simulations and corresponding \texttt{MMHH}-time stamps.
The network learns an additional embedding for the timestamps, which it provides as input in addition to the perturbed state $\rX(\difftime)$ and the noise level $\difftime$.

Conditioning the trained flow model on coarse climate simulations $\rY$ follows the mechanism used by \citet[Eq.\,(5), Supplementary Section 6]{Schmidt2025}.
This finally yields an approximation to the posterior score
\begin{equation}
  v_{t-k:t+k}^{(\theta)} \approx \nabla_\rX \log p(\rX \mid \rY, t\hod, t\mnth).
\end{equation}
We found that including additional temporal information ($t\hod, t\mnth$) greatly facilitated the learning task.
Furthermore, conditioning the model directly on temporal information makes sampling temporally aligned predictions possible, even when the coarse external information $\rY$ are long-term aggregates that span a complete diurnal cycle or more (read: climate simulations with $\geq$ daily temporal resolution).

%TC:endignore

\end{document}